# A Dual-Dimer Method for Training Physics-Constrained Neural Networks with Minimax Architecture

Dehao Liu, Yan Wang*

*Woodruff School of Mechanical Engineering, Georgia Institute of Technology, Atlanta, GA 30332, USA*

*Corresponding author. Email: yan-wang@gatech.edu  Tel: +1-404-894-4714.

## ABSTRACT

Data sparsity is a common issue to train machine learning tools such as neural networks for engineering and scientific applications, where experiments and simulations are expensive. Recently physics-constrained neural networks (PCNNs) were developed to reduce the required amount of training data. However, the weights of different losses from data and physical constraints are adjusted empirically in PCNNs. In this paper, a new physics-constrained neural network with the minimax architecture (PCNN-MM) is proposed so that the weights of different losses can be adjusted systematically. The training of the PCNN-MM is searching the high-order saddle points of the objective function. A novel saddle point search algorithm called Dual-Dimer method is developed. It is demonstrated that the Dual-Dimer method is computationally more efficient than the gradient descent ascent method for nonconvex-nonconcave functions and provides additional eigenvalue information to verify search results. A heat transfer example also shows that the convergence of PCNN-MMs is faster than that of traditional PCNNs.

# 1 INTRODUCTION

Machine learning (ML) models such as neural networks and deep learning models have been applied successfully in diverse fields. Nevertheless, data sparsity is still the main challenge to apply these models to solve complex scientific and engineering problems. The root cause is the "curse of dimensionality" in training these models. Training algorithms need to explore and exploit in a very high dimensional parameter space to search the optimal parameters for complex models. When the dimension increases, the required amount of training data grows exponentially in order to cover the space and ensure the convergence of training. Because data acquisitions in scientific experiments and high-fidelity engineering simulations are very costly, it is difficult to collect enough training data to fully train complex models. Predictions from those models will not be reliable because of overfitting.

Recently, physics-constrained machine learning emerged as a promising approach to alleviate the issue of data sparsity. In this approach, prior knowledge in science and engineering is incorporated as constraints to guide the training of ML models. In the training of physics-constrained neural networks (PCNNs) (Dissanayake & Phan‐Thien, 1994; Jianyu, Siwei, Yingjian, & Yaping, 2003; Liu & Wang, 2019; Mai-Duy & Tran-Cong, 2001; Raissi, Perdikaris, & Karniadakis, 2019; Souza De Cursi & Koscianski, 2007; Zhu, Zabaras, Koutsourelakis, & Perdikaris, 2019), physical models serve as the constraints and regularize the training loss. It has been shown that the required amount of training data can be reduced by adding physical constraints as the regularization terms. However, the training efficiency is sensitively dependent on the weights associated with the different losses with respect to data and physical constraints. In existing PCNNs, the weights were either fixed or adjusted empirically. Systematic approaches for weight adjustment are needed.

In this work, we propose a new formulation of PCNN to systematically search the optimal weights of different losses. The training of the PCNN is formulated as a minimax problem instead of minimization.

The PCNN with the minimax architecture is called PCNN-MM. The training of the PCNN-MM is searching the high-order saddle points of the objective function. The order of saddle points indicates the number of negative eigenvalues of the Hessian matrix. Most of the existing saddle point search algorithms only find first-order saddle points. The traditional gradient descent ascent (GDA) algorithm for high-order saddle points has the convergence issue for nonconvex-nonconcave functions, where the functions are neither convex in the subspace for minimization nor concave in the subspace for maximization. We also propose a novel saddle point search algorithm called Dual-Dimer method to search high-order saddle points during the training of the PCNN-MM. Two major contributions of this study include the new PCNN-MM formulation to systematically train physics-constrained neural networks and the Dual-Dimer algorithm to search high-order saddle points of nonconvex-nonconcave functions.

In the remainder of this paper, the state of the art of physics-constrained machine learning will be reviewed in Section 2. The background of our previous work (Liu & Wang, 2019) in the training of PCNNs will be introduced. Existing saddle point search methods will also be reviewed. In Section 3, the proposed PCNN-MM formulation and the Dual-Dimer algorithm will be described. The local convergence analysis of the Dual-Dimer algorithm is also included. In Section 4, the proposed Dual-Dimer algorithm is evaluated using three nonconvex-nonconcave analytical functions, including a four-dimensional (4D) Rastrigin function, a 4D Ackley function, and a 20D Styblinski–Tang function. In Section 5, a heat transfer problem is used to demonstrate the effectiveness of the Dual-Dimer algorithm, where the evolution of the 2D temperature distribution is predicted. The performance of the PCNN-MM trained by the Dual-Dimer method is compared with the PCNN with the adaptive weighting scheme and the PCNN-MM trained by the GDA method. The convergence speed and stability of different models are also tested.

## 2    BACKGROUND

The background of physics-constrained machine learning is provided in Section 2.1. Our previous work (Liu & Wang, 2019) in the training of PCNNs with the adaptive weighting scheme is introduced in Section 2.2. The training of the proposed PCNN-MM is to find the high-order saddle points of the loss function. The existing saddle point search methods are reviewed in Section 2.3.

### 2.1    Physics-Constrained Machine Learning

The basic idea of physics-constrained machine learning is to incorporate prior knowledge into ML models as constraints so that they can guide the training process. For example, the prior knowledge of the architecture and connection weights was incorporated into a neural network as constraints to improve the training efficiency (Han & Huang, 2008). The prior knowledge of functions and their derivatives was embedded into support vector regression as constraints to reduce the approximation error (Lauer & Bloch, 2008). Analytical relationships were also incorporated as the penalty terms in the objective function of neural networks to improve the prediction capability (Jia et al., 2019; H. P. N. Nagarajan et al., 2019; Read et al., 2019).

Neural networks have been used as surrogate models to approximate the solutions of ordinary differential equations (ODEs) or partial differential equations (PDEs) with reduced computational time. It was shown that neural networks such as multi-layer perceptron (MLP) and radial basis function (RBF) neural networks can solve ODEs and PDEs with higher accuracy and lower memory requirement than traditional numerical methods (Shirvany, Hayati, & Moradian, 2009). The prior knowledge of initial and boundary conditions can be incorporated in the trial solutions to improve the training efficiency of neural networks (I. E. Lagaris, Likas, & Fotiadis, 1998; Shekari Beidokhti & Malek, 2009). However, it may be difficult to find trial solutions for boundary value problems which are defined on irregular boundaries. To solve this problem, a MLP-RBF synergy model (Isaac Elias Lagaris, Likas, & Papageorgiou, 2000) was developed, where the first part of the trial solution was replaced by the RBF neural network so that the boundary conditions on irregular boundaries can be satisfied. In addition, in the constrained backpropagation training (Di Muro & Ferrari, 2008; Ferrari & Jensenius, 2008; S. He, Reif, & Unbehauen, 2000; Rudd, Muro, & Ferrari, 2014), the prior knowledge of boundary conditions was explicitly embedded as equality constraints and imposed on the weights of neural networks. Moreover, the prior knowledge of model forms and boundary conditions can be embedded as regularization terms in the loss function of a neural network to solve ODEs (Bellamine, Almansoori, & Elkamel, 2015; Malek & Shekari Beidokhti, 2006). The prior knowledge can also be embedded as regularization terms after transforming the original PDEs into their weighted residual forms (Dissanayake & Phan‐Thien, 1994). Similarly, the original model forms and boundary conditions can be directly incorporated as regularization terms into PCNNs (Jianyu et al., 2003; Mai-Duy & Tran-Cong, 2001; Raissi et al., 2019; Zhu et al., 2019). Regularization parameters can be introduced to control the trade-off between data fitting and physics-based regularization (Souza De Cursi & Koscianski, 2007).

The effectiveness of PCNNs has been demonstrated in the above work. The training of PCNNs was formulated as the minimization of a hybrid cost or loss function. The relative importance of training data and prior knowledge are adjusted by changing the weights of different losses. The drawback of this training scheme is that the weights of different losses are fixed or empirically determined, which affects the training efficiency.

## 2.2   Physics-Constrained Neural Network (PCNN) with Adaptive Weighting Scheme

The training of PCNNs with the adaptive weighting scheme (Liu & Wang, 2019) can improve the training efficiency. The scheme is introduced as follows. Suppose that a time-dependent parametric PDE is given by

$$\boldsymbol{D}[u(t,\mathbf{x})] = P\left(u, \frac{\partial u}{\partial t}, \frac{\partial u}{\partial \mathbf{x}}, \frac{\partial^2 u}{\partial t^2}, \frac{\partial^2 u}{\partial \mathbf{x}^2}, \dots\right) = f(t,\mathbf{x}), \ t \in [0,T], \ \mathbf{x} \in \Omega, \tag{1}$$

where $D[\cdot]$ is the differential operator, $u(t, \mathbf{x})$ is the true solution to be found, $f(t, \mathbf{x})$ is a source or sink term, $t$ is the time, $\mathbf{x} = (x_1, x_2, \ldots, x_n)$ is the spatial vector, and $\Omega \in \mathbb{R}^n$ denotes the definition domain. This general PDE is subject to initial conditions

$$\Lambda[u(0, \mathbf{x})] = g(\mathbf{x}) \tag{2}$$

and boundary conditions

$$\Gamma[u(t, \mathbf{x}_s)] = h(t, \mathbf{x}_s), \ t \in [0, T], \ \mathbf{x}_s \in \partial\Omega, \tag{3}$$

where $\Lambda[\cdot]$ and $\Gamma[\cdot]$ are also differential operators, and $\partial\Omega$ is the boundary of the definition domain.

The PCNN with a multilayer perceptron structure can approximate the true solution $u(t, \mathbf{x})$. The network includes one input layer $(t, \mathbf{x})$, multiple hidden layers, and one output layer $U(t, \mathbf{x})$. The weights $\mathbf{w}$ of the PCNN can be trained by minimizing the mean squared loss or total cost function (Liu & Wang, 2019)

$$\min_{\mathbf{w}} E(\mathbf{w}) = \lambda_T E_T(\mathbf{w}) + \lambda_P E_P(\mathbf{w}) + \lambda_I E_I(\mathbf{w}) + \lambda_S E_S(\mathbf{w}), \tag{4}$$

where $E_T$ the loss caused by the discrepancy between the training data and the PCNN prediction, $E_P$, $E_I$, and $E_S$ are the losses due to the violations of the model, initial conditions, and boundary conditions as specified by Eqs. (1)-(3) respectively. The weights of different losses $\lambda_T$, $\lambda_P$, $\lambda_I$, and $\lambda_S$ also satisfy constraint $\lambda_T + \lambda_P + \lambda_I + \lambda_S = 1$.

The adaptive scheme is to assign the weights of different losses as

$$\lambda_i = \frac{E_i}{E_T + E_P + E_I + E_S}, i \in \{T, P, I, S\} \tag{5}$$

for each iteration of the training process. That is, the weights are proportional to the individual losses respectively. It has been demonstrated that the adaptive weighting scheme helps improve the training efficiency of a PCNN. However, this adaptive weighting scheme is still empirical. The proposed new minimax architecture enables systematic weight adjustment.

## 2.3  Saddle Point Search Methods

The training of our new PCNN-MM is searching high-order saddle points. Various saddle point search algorithms have been developed (Alhat, Lasrado, & Wang, 2008). These include surface walking algorithm (Simons, Jørgensen, Taylor, & Ozment, 1983), DHS method (Dewar, Healy, & Stewart, 1984), partitioned rational function optimization method (Banerjee, Adams, Simons, & Shepard, 1985), activation-relaxation technique (Mousseau & Barkema, 1998), dimer method (Henkelman & Jónsson, 1999; Heyden, Bell, & Keil, 2005; Kästner & Sherwood, 2008), nudged elastic band (Henkelman & Jónsson, 2000; Henkelman, Uberuaga, & Jónsson, 2000), and curve swarm method (L. He & Wang, 2013, 2015; Tran, He, & Wang, 2018; Tran, Liu, He-Bitoun, & Wang, 2020). However, these methods can only identify first-order saddle points instead of high-order ones.

The well-known GDA algorithm has been widely used to search saddle points. In the past decade, the GDA algorithm has been applied to solve the nonconvex-nonconcave minimax problems, which arise from game theory (Leyton-Brown & Shoham, 2008), generative adversarial networks (Goodfellow et al., 2014), and robust optimization (Beyer & Sendhoff, 2007). However, it has difficulty to converge to the saddle points of the nonconvex-nonconcave functions (Daskalakis & Panageas, 2018). Some GDA extensions are also available. For instance, a proximally guided stochastic subgradient method (Rafique, Liu, Lin, & Yang, 2018) was proposed to solve a class of weakly-convex-concave minimax problems. A multi-step GDA algorithm (Nouiehed, Sanjabi, Huang, Lee, & Razaviyayn, 2019) and a proximal dual implicit accelerated gradient algorithm (Thekumparampil, Jain, Netrapalli, & Oh, 2019) were developed to solve the nonconvex but concave minimax problems. Two-time-scale GDA (Heusel, Ramsauer, Unterthiner, Nessler, & Hochreiter, 2017) was shown to converge to stationary local Nash equilibria under certain strong conditions. Symplectic gradient adjustment (SGA) algorithm (Balduzzi et al., 2018) was proposed to search stable fixed points in general games, including potential games and Hamiltonian games. Hessian-based algorithms (Adolphs, Daneshmand, Lucchi, & Hofmann, 2018; Mazumdar, Jordan, & Sastry, 2019) were developed to search local saddle points in the nonconvex-nonconcave settings. However, the computation of the Hessian matrix is expensive for high-dimensional problems.

## 3    METHODOLOGY

Here, we propose a new generic formulation of physics-constrained neural networks with the minimax architecture. The adjustment of weights associated with physical constraints can be done systematically during the training process. A new high-order saddle point search method is also developed to train the new PCNNs with nonconvex-nonconcave objective functions. The formulation of the PCNN-MM is described in Section 3.1. The generic Dual-Dimer saddle point search method is introduced in Section 3.2.

### 3.1    Physics-Constrained Neural Network with Minimax Architecture (PCNN-MM)

The training of the PCNN-MM is to solve the minimax problem

$$\min_{\mathbf{w}} \max_{\boldsymbol{\alpha}} E(\mathbf{w}, \boldsymbol{\alpha}) = \lambda_T(\boldsymbol{\alpha})E_T(\mathbf{w}) + \lambda_P(\boldsymbol{\alpha})E_P(\mathbf{w}) + \lambda_I(\boldsymbol{\alpha})E_I(\mathbf{w}) + \lambda_S(\boldsymbol{\alpha})E_S(\mathbf{w}), \tag{6}$$

where the weights of different losses $\lambda_T$, $\lambda_P$, $\lambda_I$, and $\lambda_S$ are now functions of parameters $\boldsymbol{\alpha} = (\alpha_T, \alpha_P, \alpha_I, \alpha_S)$. The formulation in Eq. (6) can be regarded as a generalization of the formulation in Eq. (4). Training is to minimize the possible loss for a worst-case (maximum loss) scenario. That is, we perform the maximization of the total loss $E(\mathbf{w}, \boldsymbol{\alpha})$ over the parameter subspace of $\boldsymbol{\alpha}$ and the minimization of the total loss over the parameter subspace of $\mathbf{w}$. During the training of the PCNN-MM, the weights of different losses $\lambda$'s will be adjusted to maximize the total loss $E(\mathbf{w}, \boldsymbol{\alpha})$ in $\boldsymbol{\alpha}$ subspace, whereas the weights of the neural network $\mathbf{w}$'s will be tuned to minimize the total loss $E(\mathbf{w}, \boldsymbol{\alpha})$. When one of the losses is larger than the other ones, its corresponding weight tends to increase to emphasize the importance of that particular loss so that the total loss is maximized. To counteract, the weights of the neural network will be adjusted to minimize the total loss so that the total loss can be reduced faster. That is how the weights of different losses are systematically adjusted. In this work, the weights of different losses are defined as the softmax functions as

$$\lambda_i(\boldsymbol{\alpha}) = \frac{\exp(\alpha_i)}{\exp(\alpha_T) + \exp(\alpha_P) + \exp(\alpha_I) + \exp(\alpha_S)}, i \in \{T, P, I, S\}. \tag{7}$$

After applying softmax functions, the range of the weights of different losses $\lambda_i$ will be in the interval $[0,1]$, and they will add up to one.

Let $\boldsymbol{\theta} = (\mathbf{w}, \boldsymbol{\alpha})$ denote the optimization parameters for objective function $E$. The training of the PCNN-MM is to find a minimax point or saddle point on a high-dimensional energy landscape $E$. The training of the PCNN-MM, which is to solve the minimax problem in Eq. (6), is equivalent to finding a saddle point $\boldsymbol{\theta}^* = (\mathbf{w}^*, \boldsymbol{\alpha}^*)$ such that

$$E(\mathbf{w}^*, \boldsymbol{\alpha}) \leq E(\mathbf{w}^*, \boldsymbol{\alpha}^*) \leq E(\mathbf{w}, \boldsymbol{\alpha}^*) \ (\forall \mathbf{w} \in \mathbb{R}^D, \forall \boldsymbol{\alpha} \in \mathbb{R}^4). \tag{8}$$

That is, the saddle point is the minimum in $\mathbf{w}$ subspace and maximum in $\boldsymbol{\alpha}$ subspace. The sufficient conditions for $\boldsymbol{\theta}^* = (\mathbf{w}^*, \boldsymbol{\alpha}^*)$ to be the desired saddle point are: (1) the gradients of the objective function with respect to $(\mathbf{w}, \boldsymbol{\alpha})$ are zeros, i.e., $\nabla_{\mathbf{w}} E(\boldsymbol{\theta}^*) = \mathbf{0}$ and $\nabla_{\boldsymbol{\alpha}} E(\boldsymbol{\theta}^*) = \mathbf{0}$; (2) the second derivatives $\nabla_{\mathbf{w}}^2 E(\boldsymbol{\theta}^*)$ in the $\mathbf{w}$ subspace are positive semi-definite; and (3) the second derivatives $\nabla_{\boldsymbol{\alpha}}^2 E(\boldsymbol{\theta}^*)$ in the $\boldsymbol{\alpha}$ subspace are negative semi-definite.

## 3.2 The Dual-Dimer Method

It is known that the steepest step $\Delta\boldsymbol{\theta}$ to reach a stationary point (local minimum, local maximum, or saddle point) can be obtained by Newton's method

$$\Delta\boldsymbol{\theta} = \mathbf{H}^{-1}\mathbf{f} = \sum_i \frac{(\mathbf{v}_i \cdot \mathbf{f})\mathbf{v}_i}{\beta_i}, \tag{9}$$

where $\mathbf{f} = -\nabla E$ is the force, $\mathbf{H}$ is the Hessian matrix, $\mathbf{v}_i$ is the eigenvector, and $\beta_i$ is the corresponding eigenvalue. The drawback of the gradient descent method is not the search direction but the size of the step along each eigenvector direction. Therefore, a small step should be taken along the direction $\mathbf{v}_i$ when the corresponding eigenvalue $\beta_l$ is small. By rescaling the gradients in each direction with the inverse of the corresponding eigenvalue, the Newton's method in Eq.(9) can accelerate the convergence. However, in high-dimensional problems, the computations of all eigenvectors and eigenvalues are very expensive.

The Dual-Dimer method is designed to improve the computational efficiency for high-dimensional problems. Let $\beta_s$ denotes the minimum eigenvalue of $\nabla_{\mathbf{w}}^2 E(\boldsymbol{\theta})$ with its corresponding eigenvector $\mathbf{v}_s$, and $\beta_l$ denotes the maximum eigenvalue of $\nabla_{\boldsymbol{\alpha}}^2 E(\boldsymbol{\theta})$ with its corresponding eigenvector $\mathbf{v}_l$. By augmenting the gradient descent ascent with the rescaled projections of the force along the extreme eigenvectors $(\mathbf{v}_s, \mathbf{v}_l)$, the step to reach the desired high-order saddle point in the Dual-Dimer method is given by

$$\Delta\boldsymbol{\theta} = (\Delta\boldsymbol{\theta_w}, \Delta\boldsymbol{\theta_\alpha}) + (\Delta\boldsymbol{\theta}_s, \Delta\boldsymbol{\theta}_l) = \eta\left(-\nabla_{\mathbf{w}}E(\boldsymbol{\theta}), \nabla_{\boldsymbol{\alpha}}E(\boldsymbol{\theta})\right) + \left(-\frac{(\mathbf{v}_s \cdot \nabla_{\mathbf{w}}E(\boldsymbol{\theta}))\mathbf{v}_s}{|\beta_s|}, \frac{(\mathbf{v}_l \cdot \nabla_{\boldsymbol{\alpha}}E(\boldsymbol{\theta}))\mathbf{v}_l}{|\beta_l|}\right), \quad (10)$$

where $\Delta\boldsymbol{\theta_w}$ is the gradient descent sub-step given by the first-order gradient-based optimization method (Kingma & Ba, 2014) in the $\mathbf{w}$ subspace, and $\Delta\boldsymbol{\theta_\alpha}$ is the gradient ascent sub-step in the $\boldsymbol{\alpha}$ subspace. $\eta$ is the learning rate for the gradient descent ascent sub-steps. $\Delta\boldsymbol{\theta}_s$ is the projection of the force along the $\mathbf{v}_s$ direction, and $\Delta\boldsymbol{\theta}_l$ is the projection of the force along the $\mathbf{v}_l$ direction. With augmented sub-steps $\Delta\boldsymbol{\theta}_s$ and $\Delta\boldsymbol{\theta}_l$, it is expected that at the end of the training $\nabla_{\mathbf{w}}^2 E(\boldsymbol{\theta}^*)$ does not have negative eigenvalues in $\mathbf{w}$ and $\nabla_{\boldsymbol{\alpha}}^2 E(\boldsymbol{\theta}^*)$ does not have positive eigenvalues in $\boldsymbol{\alpha}$. Therefore, the use of the extreme eigenvalues and eigenvectors in the Dual-Dimer method is to make sure that the high-order saddle points are found.

In the original dimer method (Henkelman & Jónsson, 1999; Heyden et al., 2005; Kästner & Sherwood, 2008), a dimer is rotated to find the minimum curvature direction and then translated to a first-order saddle point. The minimum curvature direction corresponds to the extreme eigenvector in the minimum subspace for the first-order saddle point. In the proposed Dual-Dimer method, the way to calculate extreme eigenvalues and eigenvectors for first-order saddle points in the original dimer method is adopted and extended to calculate the extreme values in both the minimum and maximum subspaces for high-order saddle points. The proposed Dual-Dimer method is also different from the dimer method by rescaling the step sizes along the extreme eigenvectors with the inverse of the extreme eigenvalues. The extreme eigenvalues $(\beta_s, \beta_l)$ and eigenvectors $(\mathbf{v}_s, \mathbf{v}_l)$ are computed by rotating two dimers in the subspaces of $\mathbf{w}$ and $\boldsymbol{\alpha}$ without expensive calculations of the Hessian matrix $\mathbf{H}$. The first dimer in the $\mathbf{w}$ subspace is composed of two endpoints $\boldsymbol{\theta}_1$ and $\boldsymbol{\theta}_2$, which are slightly displaced by the fixed dimer length $2\Delta R$. The locations of the endpoints $\boldsymbol{\theta}_1$ and $\boldsymbol{\theta}_2$ are given by

$$\begin{cases} \boldsymbol{\theta}_1 = \boldsymbol{\theta}_0 + \Delta R \mathbf{n} \\ \boldsymbol{\theta}_2 = \boldsymbol{\theta}_0 - \Delta R \mathbf{n} \end{cases}, \tag{11}$$

where $\mathbf{n}$ is the unit vector along the dimer axis and $\boldsymbol{\theta}_0$ is the midpoint of the dimer. Here, the components of $\mathbf{n}$ in the $\mathbf{w}$ subspace are nonzero, whereas the components of $\mathbf{n}$ in the $\boldsymbol{\alpha}$ subspace are always zero. Therefore, the rotation of the first dimer is confined in the $\mathbf{w}$ subspace. The dimer axis $\mathbf{n}$ is rotated into the smallest curvature direction of the potential energy $C(\mathbf{n})$ at the dimer midpoint $\boldsymbol{\theta}_0$, which is to solve the minimization problem

$$\min_{\mathbf{n}} C(\mathbf{n}) = \mathbf{n}^T \mathbf{H} \mathbf{n} \approx \frac{(\mathbf{f}_2 - \mathbf{f}_1) \cdot \mathbf{n}}{2\Delta R}, \tag{12}$$

where $\mathbf{H}$ is the Hessian matrix at the dimer midpoint $\boldsymbol{\theta}_0$. $\mathbf{f}_1 = -\nabla E(\boldsymbol{\theta}_1)$ and $\mathbf{f}_2 = -\nabla E(\boldsymbol{\theta}_2)$ are the forces at the locations $\boldsymbol{\theta}_1$ and $\boldsymbol{\theta}_2$, respectively. It is noted that only first derivatives are required to estimate curvatures in Eq.(12). This is the reason that the Dual-Dimer method is computationally efficient. Furthermore, the curvature $C(\mathbf{n})$ becomes the eigenvalue if $\mathbf{n}$ is the eigenvector of the Hessian matrix. Once the smallest curvature $C(\mathbf{n})$ is found, the minimum eigenvalue $\beta_s$ in the $\mathbf{w}$ subspace is equal to $C(\mathbf{n})$ and the components of $\mathbf{n}$ in the $\mathbf{w}$ subspace becomes the extreme eigenvector $\mathbf{v}_s$. The minimization problem in Eq. (12) is numerically solved by rotating the dimer. The details can be found in the original dimer method (Henkelman & Jónsson, 1999; Heyden et al., 2005; Kästner & Sherwood, 2008).

Similarly, the second dimer in the $\boldsymbol{\alpha}$ subspace is composed of two endpoints $\boldsymbol{\theta}_3$ and $\boldsymbol{\theta}_4$ with their locations given by

$$\begin{cases} \boldsymbol{\theta}_3 = \boldsymbol{\theta}_0 + \Delta R \mathbf{m} \\ \boldsymbol{\theta}_4 = \boldsymbol{\theta}_0 - \Delta R \mathbf{m} \end{cases}, \tag{13}$$

where $\mathbf{m}$ is the unit vector along the dimer axis. Here, the components of $\mathbf{m}$ in the $\boldsymbol{\alpha}$ subspace are nonzero, whereas the components of $\mathbf{m}$ in the $\mathbf{w}$ subspace are always zero. Therefore, the rotation of the second dimer is confined in the $\boldsymbol{\alpha}$ subspace. The dimer axis $\mathbf{m}$ is rotated into the largest curvature direction of the potential energy, which is to solve the maximization problem

$$\max_{\mathbf{m}} C(\mathbf{m}) = \mathbf{m}^T \mathbf{H} \mathbf{m} \approx \frac{(\mathbf{f}_4 - \mathbf{f}_3) \cdot \mathbf{m}}{2\Delta R}, \tag{14}$$

where $\mathbf{f}_3 = -\nabla E(\boldsymbol{\theta}_3)$ and $\mathbf{f}_4 = -\nabla E(\boldsymbol{\theta}_4)$ are the forces at the locations $\boldsymbol{\theta}_3$ and $\boldsymbol{\theta}_4$, respectively. Once the largest curvature $C(\mathbf{m})$ is found, the maximum eigenvalue $\beta_l$ in the $\boldsymbol{\alpha}$ subspace is equal to $C(\mathbf{m})$ and the components of $\mathbf{m}$ in the $\boldsymbol{\alpha}$ subspace become the extreme eigenvector $\mathbf{v}_l$.

The algorithm of the Dual-Dimer method is shown in Table 1. Iteratively, the sub-steps $\Delta\boldsymbol{\theta}_\mathbf{w}$, $\Delta\boldsymbol{\theta}_{\boldsymbol{\alpha}}$, $\Delta\boldsymbol{\theta}_s$, and $\Delta\boldsymbol{\theta}_l$ are calculated and the estimate saddle point location is updated. There are five hyperparameters $(m, \delta, \gamma, \eta, \varepsilon)$ that need to be tuned in the Dual-Dimer method. Parameter $m$ represents the frequency of updating extreme eigenvalues and eigenvectors. If $m$ is small, the overall computational cost will be high. If $m$ is large, the estimations of current extreme eigenvalues and eigenvectors are not accurate. Parameter $\delta$ is introduced in the algorithm to avoid the zero-division error. When the eigenvalue is close to zero, it means that the curvature is very small and the saddle point degenerates. Parameter $\gamma$ means the maximum step length of $\Delta\boldsymbol{\theta}_s$ and $\Delta\boldsymbol{\theta}_l$ to make sure that the training is converged. Parameter $\eta$ is the learning rate for the gradient descent ascent sub-steps. If $\eta$ is small, the training will be slow. If $\eta$ is large, the training may be unstable. When the objective function $E$ or the norm of the force $\|\mathbf{f}\|_2$ is less than the threshold $\varepsilon$, the search for the saddle points stops. Trade-offs need to be made between the computational accuracy and efficiency for these hyperparameters to improve the overall performance of the Dual-Dimer method. Sensitivity studies were done in this work to tune them. A more systematic method to find the optimal hyperparameters is needed in future work.

Table 1. The Dual-Dimer algorithm

| Input: | initial optimization parameters $\boldsymbol{\theta}_0 = (\mathbf{w}_0, \boldsymbol{\alpha}_0)$, objective function $E$, hyperparameters $m$, $\delta, \gamma, \eta, \varepsilon$. |
|---|---|
| Output: | desired saddle point $\boldsymbol{\theta}^*$ |
| Procedure: | 1. Initialize the iteration $t = 0$, $\boldsymbol{\theta}_t = \boldsymbol{\theta}_0$ |
| | 2. Evaluate energy $E(\boldsymbol{\theta}_t)$ and force $\mathbf{f} = -\nabla E$ |
| | 3. When $t \bmod m = 0$, compute the extreme eigenvalues $(\beta_s, \beta_l)$ and eigenvectors $(\mathbf{v}_s, \mathbf{v}_l)$ by rotating two dimers in the subspaces of $\mathbf{w}$ and $\boldsymbol{\alpha}$ |
| | 4. Calculate $\Delta\boldsymbol{\theta}_\mathbf{w} = -\eta\nabla_\mathbf{w} E(\boldsymbol{\theta})$ and $\Delta\boldsymbol{\theta}_\alpha = \eta\nabla_\alpha E(\boldsymbol{\theta})$ |
| | 5. If $\|\beta_s\| > \delta$, $\Delta\boldsymbol{\theta}_s = -\frac{(\mathbf{v}_s \cdot \nabla_\mathbf{w} E(\boldsymbol{\theta}))\mathbf{v}_s}{\|\beta_s\|}$; otherwise, $\Delta\boldsymbol{\theta}_s = \mathbf{0}$; If $\|\beta_l\| > \delta$, $\Delta\boldsymbol{\theta}_l = \frac{(\mathbf{v}_l \cdot \nabla_\alpha E(\boldsymbol{\theta}))\mathbf{v}_l}{\|\beta_l\|}$; otherwise, $\Delta\boldsymbol{\theta}_l = \mathbf{0}$ |
| | 6. If $\|\Delta\boldsymbol{\theta}_s\|_2 > \gamma$, $\Delta\boldsymbol{\theta}_s = \gamma\frac{\Delta\boldsymbol{\theta}_s}{\|\Delta\boldsymbol{\theta}_s\|_2}$; If $\|\Delta\boldsymbol{\theta}_l\|_2 > \gamma$, $\Delta\boldsymbol{\theta}_l = \gamma\frac{\Delta\boldsymbol{\theta}_l}{\|\Delta\boldsymbol{\theta}_l\|_2}$ |
| | 7. $t = t + 1$ |
| | 8. Update optimization parameters by calculating $\Delta\boldsymbol{\theta} = (\Delta\boldsymbol{\theta}_\mathbf{w}, \Delta\boldsymbol{\theta}_\alpha) + (\Delta\boldsymbol{\theta}_s, \Delta\boldsymbol{\theta}_l)$ and $\boldsymbol{\theta}_t = \boldsymbol{\theta}_{t-1} + \Delta\boldsymbol{\theta}$ |
| | 9. Return to step 2 until $\|\mathbf{f}\|_2 < \varepsilon$ or $E < \varepsilon$ |
| | 10. Output $\boldsymbol{\theta}^* = \boldsymbol{\theta}_t$ |

## 3.3 Local Convergence

The local convergence of the Dual-Dimer method is analyzed here. Let us define a fixed-point function

$$F(\boldsymbol{\theta}) = \boldsymbol{\theta} + \eta\big(-\nabla_\mathbf{w} E(\boldsymbol{\theta}), \nabla_\alpha E(\boldsymbol{\theta})\big) + \left(-\frac{(\mathbf{v}_s \cdot \nabla_\mathbf{w} E(\boldsymbol{\theta}))\mathbf{v}_s}{\|\beta_s\|}, \frac{(\mathbf{v}_l \cdot \nabla_\alpha E(\boldsymbol{\theta}))\mathbf{v}_l}{\|\beta_l\|}\right) \quad (15)$$

and assume that $F(\boldsymbol{\theta})$ is differentiable. The desired saddle point $\boldsymbol{\theta}^*$ can be found by iteratively applying the fixed-point function $F(\boldsymbol{\theta})$. If $\beta_s = 0$ and $\beta_l = 0$, as shown in Table 1, then the fixed-point iteration becomes the GDA method, which is locally stable according to (Mescheder, Nowozin, & Geiger, 2017; V. Nagarajan & Kolter, 2017). If $\beta_s \neq 0$ and $\beta_l \neq 0$, we have the following lemmas and theorem. The proofs can be found in the appendix.

**Lemma 1.** The Jacobian of the loss function at the desired saddle point $\boldsymbol{\theta}^* = (\mathbf{w}^*, \boldsymbol{\alpha}^*)$ is

$$\nabla F(\boldsymbol{\theta}^*) = \mathbf{I} + \eta\begin{pmatrix} -\nabla_\mathbf{w}^2 E(\boldsymbol{\theta}^*) & -\nabla_{\mathbf{w},\alpha}^2 E(\boldsymbol{\theta}^*) \\ \nabla_{\alpha,\mathbf{w}}^2 E(\boldsymbol{\theta}^*) & \nabla_\alpha^2 E(\boldsymbol{\theta}^*) \end{pmatrix} + \begin{pmatrix} -\frac{1}{\beta_s}\mathbf{v}_s\mathbf{v}_s^T\nabla_\mathbf{w}^2 E(\boldsymbol{\theta}^*) & -\frac{1}{\beta_s}\mathbf{v}_s\mathbf{v}_s^T\nabla_{\mathbf{w},\alpha}^2 E(\boldsymbol{\theta}^*) \\ -\frac{1}{\beta_l}\mathbf{v}_l\mathbf{v}_l^T\nabla_{\alpha,\mathbf{w}}^2 E(\boldsymbol{\theta}^*) & -\frac{1}{\beta_l}\mathbf{v}_l\mathbf{v}_l^T\nabla_\alpha^2 E(\boldsymbol{\theta}^*) \end{pmatrix}, (16)$$

where $\mathbf{I}$ is the real-valued identity matrix. If there exists an $\eta$ ($\eta > 0$) such that the absolute values of all the eigenvalues of $\nabla F(\boldsymbol{\theta}^*)$ are less than 1, then there is an open neighborhood $K$ of $\boldsymbol{\theta}^*$ so that for all $\boldsymbol{\theta} \in K$, the fixed-point iterations of $F(\boldsymbol{\theta})$ in Eq. (15) are stable in $K$. The rate of convergence is at least linear.

**Lemma 2.** Let $\beta_A = a + bi$ be the eigenvalues of the matrix $\mathbf{A}$, $\beta_B = c + di$ be the eigenvalues of the matrix $\mathbf{B}$, where $i = \sqrt{-1}$. The eigenvalues of the matrix $\mathbf{I} + \eta\mathbf{A} + \mathbf{B}$, where $\eta > 0$, lie in the unit ball if

$$\Delta = [2(a + ac + bd)]^2 - 4(a^2 + b^2)(c^2 + 2c + d^2) > 0 \tag{17}$$

and

$$\begin{cases} 0 < \eta < \frac{-2(a+ac+bd)+\sqrt{\Delta}}{2(a^2+b^2)}, \; if \; a + ac + bd \geq 0 \; and \; c^2 + 2c + d^2 > 0 \\ max\left\{0, \frac{-2(a+ac+bd)-\sqrt{\Delta}}{2(a^2+b^2)}\right\} < \eta < \frac{-2(a+ac+bd)+\sqrt{\Delta}}{2(a^2+b^2)}, \; if \; a + ac + bd < 0 \end{cases} \tag{18}$$

for all eigenvalues $\beta_A$ of $\mathbf{A}$ and $\beta_B$ of $\mathbf{B}$.

**Theorem 1.** Let $\boldsymbol{\theta}^* = (\mathbf{w}^*, \boldsymbol{\alpha}^*)$ be the desired saddle point, $\beta_A = a + bi$ be the eigenvalues of $\mathbf{A} = \begin{pmatrix} -\nabla_{\mathbf{w}}^2 E(\boldsymbol{\theta}^*) & -\nabla_{\mathbf{w},\boldsymbol{\alpha}}^2 E(\boldsymbol{\theta}^*) \\ \nabla_{\boldsymbol{\alpha},\mathbf{w}}^2 E(\boldsymbol{\theta}^*) & \nabla_{\boldsymbol{\alpha}}^2 E(\boldsymbol{\theta}^*) \end{pmatrix}$, $\beta_B = c + di$ be the eigenvalues of $\mathbf{B} = \begin{pmatrix} -\frac{1}{\beta_s}\mathbf{v}_s\mathbf{v}_s^T\nabla_{\mathbf{w}}^2 E(\boldsymbol{\theta}^*) & -\frac{1}{\beta_s}\mathbf{v}_s\mathbf{v}_s^T\nabla_{\mathbf{w},\boldsymbol{\alpha}}^2 E(\boldsymbol{\theta}^*) \\ -\frac{1}{\beta_l}\mathbf{v}_l\mathbf{v}_l^T\nabla_{\boldsymbol{\alpha},\mathbf{w}}^2 E(\boldsymbol{\theta}^*) & -\frac{1}{\beta_l}\mathbf{v}_l\mathbf{v}_l^T\nabla_{\boldsymbol{\alpha}}^2 E(\boldsymbol{\theta}^*) \end{pmatrix}$, and $\eta > 0$. The fixed-point iterations of $F(\boldsymbol{\theta})$ in Eq. (15) are locally stable if

$$\Delta = [2(a + ac + bd)]^2 - 4(a^2 + b^2)(c^2 + 2c + d^2) > 0 \tag{19}$$

and

$$\begin{cases} 0 < \eta < \frac{-2(a+ac+bd)+\sqrt{\Delta}}{2(a^2+b^2)}, \; if \; a + ac + bd \geq 0 \; and \; c^2 + 2c + d^2 > 0 \\ max\left\{0, \frac{-2(a+ac+bd)-\sqrt{\Delta}}{2(a^2+b^2)}\right\} < \eta < \frac{-2(a+ac+bd)+\sqrt{\Delta}}{2(a^2+b^2)}, \; if \; a + ac + bd < 0 \end{cases} \tag{20}$$

for all eigenvalues $\beta_A$ of $\mathbf{A}$ and $\beta_B$ of $\mathbf{B}$.

## 4 EVALUATION OF THE ALGORITHM

The proposed Dual-Dimer algorithm is evaluated with three analytical nonconvex-nonconcave functions. They are a 4D Rastrigin function, a 4D Ackley function, and a 20D Styblinski–Tang function. The first saddle point problem of the 4D Rastrigin function is given by

$$\min_{x_1,x_2}\max_{x_3,x_4} E(\mathbf{x}) = \sum_{i=1}^{4}[x_i^2 - 10\cos(2\pi x_i) + 10],\tag{21}$$

The second problem of the 4D non-separable Ackley function is given by

$$\min_{x_1,x_2}\max_{x_3,x_4} E(\mathbf{x}) = -20exp\left(-0.2\sqrt{\frac{1}{4}\sum_{i=1}^{4}x_i^2}\right) - exp\left(\frac{1}{4}\sum_{i=1}^{4}\cos(2\pi x_i)\right) + 20 + e,\tag{22}$$

The third one of the 20D Styblinski–Tang function is given by

$$\min_{x_1\sim x_{10}}\max_{x_{11}\sim x_{20}} E(\mathbf{x}) = \frac{1}{2}\sum_{i=1}^{20}[x_i^4 - 16x_i^2 + 5x_i],\tag{23}$$

There are multiple stationary points on the surfaces of these analytical functions, which makes it difficult to find high-order saddle points. Since the objective functions are analytical, the gradients and Hessian matrices of the objective functions can be computed easily. Therefore, the high-order saddle points can be easily verified.

Both GDA and Dual-Dimer methods are used to search a second-order saddle point of the 4D Rastrigin function, a second-order saddle point of the 4D Ackley function, and a tenth-order saddle point of the 20D Styblinski–Tang function. The gradient descent ascent steps in the GDA and Dual-Dimer method are given by the Adam algorithm with the learning rate of $5 \times 10^{-4}$. The dimer distance is $2\Delta R = 2 \times 10^{-4}$. The hyperparameters of the Dual-Dimer method in examples of analytical functions are listed in Table 2. The search stops when the norm of the force is less than the threshold ($\|\mathbf{f}\|_2 < \varepsilon$).

Table 2. Hyperparameters of the Dual-Dimer method in examples of analytical functions

| Hyperparameters | Value |
| --- | --- |
| Frequency of updating extreme eigenvalues and eigenvectors, $m$ | 40 |
| The parameter to avoid the zero-division error, $\delta$ | $1 \times 10^{-3}$ |
| Maximum step length of $\Delta\boldsymbol{\theta}_s$ and $\Delta\boldsymbol{\theta}_l$, $\gamma$ | 0.1 |
| Learning rate for the gradient descent ascent sub-steps, $\eta$ | $5 \times 10^{-4}$ |
| The threshold for stopping search ($\|\mathbf{f}\|_2 < \varepsilon$), $\varepsilon$ | $1 \times 10^{-4}$ |

The high-order saddle points found by the GDA and Dual-Dimer methods are listed in Table 3. In the examples of Rastrigin and Ackley functions, the second-order saddle points $\mathbf{x}^*$ found by the GDA and Dual-Dimer methods are the same. In the example of Styblinski–Tang function, two different tenth-order saddle points were found by the GDA and Dual methods. By changing the random seed, different second-order saddle points can be found by the GDA and Dual-Dimer method. Since variables in the Rastrigin and Styblinski–Tang functions are separable, all off-diagonal elements of their Hessian matrices are zeros. Therefore, the diagonal elements of their Hessian matrices are eigenvalues. On the contrary, since variables in Ackley function are non-separable, some off-diagonal elements of its Hessian matrix are nonzero. It is shown in Table 3 that the extreme eigenvalues ($\beta_s, \beta_l$) calculated by the Dual-Dimer method agree well with the true extreme eigenvalues ($\beta_s^*, \beta_l^*$). It is noted that the GDA method does not provide additional eigenvalue information, whereas the Dual-Dimer method provides. It is easy to verify that the norms of the gradient $\|\nabla E(\mathbf{x}^*)\|_2$ at all identified saddle points are less than $1 \times 10^{-4}$. The minimum eigenvalue $\beta_s$ in the minimum subspace at the saddle point $\mathbf{x}^*$ is positive, whereas the maximum eigenvalue $\beta_l$ in the maximum subspace at the saddle point $\mathbf{x}^*$ is negative. It is demonstrated that the high-order saddle points of these nonconvex-nonconcave analytical functions can be found by the Dual-Dimer method.

Table 3. High-order saddle points found by the GDA and Dual-Dimer method

| | 4D Rastrigin function | 4D Ackley function | 20D Styblinski–Tang function |
|---|---|---|---|
| Saddle point $\mathbf{x}^*$ | $\begin{pmatrix} -0.9950 \\ -0.9950 \\ 0.5025 \\ 0.5025 \end{pmatrix}$ | $\begin{pmatrix} 0.9532 \\ 0 \\ -2.6489 \\ 0.5255 \end{pmatrix}$ | $x_i = \begin{cases} -2.9035 & i = 1,2,3,4,6,7,10 \\ 2.7468 & i = 5,8,9 \\ 0.1567 & i = 11{\sim}20 \end{cases}$ (GDA) $\qquad$ $x_i = \begin{cases} -2.9035 & i = 1,2,5,6 \\ 2.7468 & i = 2,3,7,8,9,10 \\ 0.1567 & i = 11{\sim}20 \end{cases}$ (Dual-Dimer) |
| True minimum eigenvalue $\beta_s^*$ in the minimum subspace | $\nabla^2_{x_i=-0.9950} E(\mathbf{x}^*)$ $= 396.53$ | 10.64 | $\nabla^2_{x_i=2.7468} E(\mathbf{x}^*) = 29.30$ |
| True maximum eigenvalue $\beta_l^*$ in the maximum subspace | $\nabla^2_{x_i=0.5025} E(\mathbf{x}^*)$ $= -392.62$ | −8.18 | $\nabla^2_{x_i=0.1567} E(\mathbf{x}^*) = -15.85$ |
| Calculated minimum eigenvalue $\beta_s$ in the minimum subspace by Dual-Dimer | 396.53 | 10.83 | 29.30 |
| Calculated maximum eigenvalue $\beta_l$ in the maximum subspace by Dual-Dimer | −392.62 | −8.13 | −15.85 |

In addition, Fig. 1 shows the changes in the forces or gradients for the two methods during the search for saddle points of the three analytical functions. It is seen that the force for the Dual-Dimer method decreases faster than the GDA method. The results show that the Dual-Dimer method is computationally more efficient than the GDA method to find these high-order saddle points. Table 4 shows the quantitative comparison of the convergence between the GDA and Dual-Dimer methods. The convergence speeds of the Dual-Dimer method are about 10 times, 9 times, and 2 times faster than those of the GDA method for the Rastrigin, Ackley, and Styblinski–Tang functions, respectively.

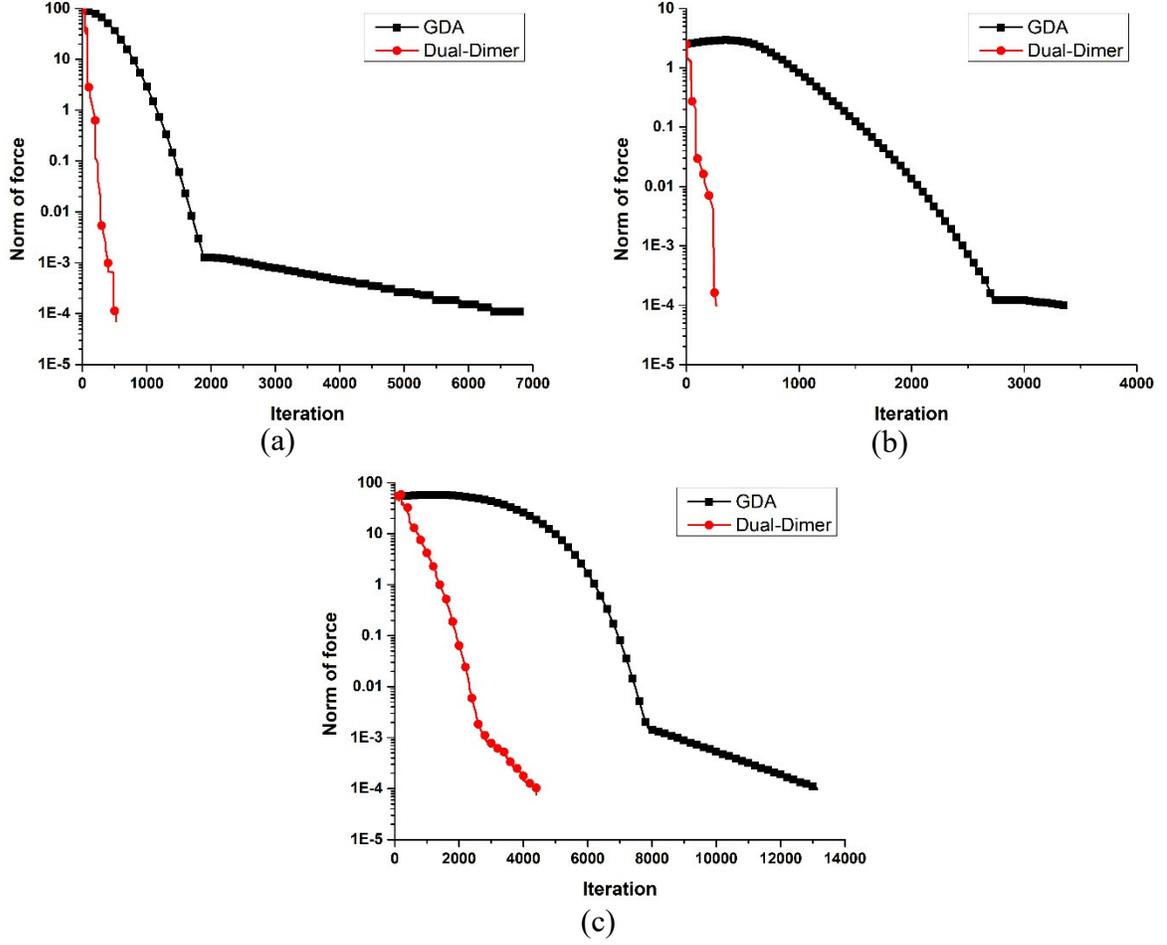

Fig. 1. The change in the force during the search for saddle points of (a) a 4D Rastrigin function, (b) a 4D Ackley function, and (c) a 20D Styblinski–Tang function.

Table 4. Comparison of convergence speeds of the GDA and Dual-Dimer methods

| Methods | 4D Rastrigin function | | 4D Ackley function | | 20D Styblinski–Tang function | |
|---|---|---|---|---|---|---|
| | Training iteration | Training time (second) | Training iteration | Training time (second) | Training iteration | Training time (second) |
| GDA | 6840 | 6.56 | 3366 | 3.23 | 13136 | 44.70 |
| Dual-Dimer | **522** | **0.58** | **265** | **0.31** | **4403** | **16.55** |

## 5    DEMONSTRATION

In this section, a heat transfer example is used to demonstrate the increased computational efficiency of PCNNs by adopting the new minimax architecture.  In the heat transfer problem, the evolution of the 2D temperature distribution is predicted by a PCNN with the adaptive weighting scheme, a PCNN-MM trained by the GDA method, and a PCNN-MM trained by the Dual-Dimer method. The PCNN setup is described in Section 5.1. The computational results and a quantitative comparison for different models are provided in Section 5.2. The convergence speed and stability of different models are also investigated.

## 5.1  PCNN Setup

In this example, the 2D heat equation with the zero Neumann boundary condition is given by

$$\begin{cases} u_t - 0.01(u_{xx} + u_{yy}) = 0, & t, x, y \in [0,1], \\ u(0, x, y) = 0.5[sin(4\pi x) + sin(4\pi y)], \\ \quad u_x(t, 0, y) = 0, \\ \quad u_x(t, 1, y) = 0, \\ \quad u_y(t, x, 0) = 0, \\ \quad u_y(t, x, 1) = 0. \end{cases} \tag{24}$$

where $u$ is the 2D temperature field.

The total loss function in a PCNN is defined by Eq. (4), whereas the total loss in a PCNN-MM is defined by Eq. (6). The training loss is

$$E_T = \frac{1}{N_T} \sum_{i=1}^{N_T} \left| U(t_i^T, x_i^T, y_i^T) - T(t_i^T, x_i^T, y_i^T) \right|^2. \tag{25}$$

The physical loss is given by

$$E_P = \frac{1}{N_P} \sum_{i=1}^{N_P} \left| U_t(t_i^P, x_i^P, y_i^P) - 0.01\left[ U_{xx}(t_i^P, x_i^P, y_i^P) + U_{yy}(t_i^P, x_i^P, y_i^P) \right] \right|^2. \tag{26}$$

The initial loss is

$$E_I = \frac{1}{N_I} \sum_{i=1}^{N_I} \left| U(0, x_i^I, y_i^I) - 0.5[sin(4\pi x_i^I) + sin(4\pi y_i^I)] \right|^2. \tag{27}$$

The boundary loss is given by

$$E_S = \frac{1}{N_S} \sum_{i=1}^{N_S} \left[ \left| U_x(t_i^S, 0, y_i^S) \right|^2 + \left| U_x(t_i^S, 1, y_i^S) \right|^2 + \left| U_y(t_i^S, x_i^S, 0) \right|^2 + \left| U_y(t_i^S, x_i^S, 1) \right|^2 \right]. \tag{28}$$

The weights of different losses in the traditional PCNN are adjusted dynamically by the adaptive weighting scheme given in Eq. (5), whereas the weights of different losses in a PCNN-MM are defined in Eq. (7).

The construction of the PCNN and PCNN-MMs is accomplished by using PyTorch (Paszke et al., 2019), which is an open-source Python library for machine learning. The PCNN and PCNN-MMs have the same structure of 30-20-30-20, where each network has 4 layers and the numbers of neurons in these layers are 30, 20, 30, and 20 respectively. The neural network architecture was identified by conducting some simple sensitivity studies. The hyperbolic tangent (tanh) function is used as the activation function.

The training data for the heat transfer example come from the finite-element method (FEM) solutions. The simulation domain is $x, y \in [0,1]$ and the time period is $t \in [0,1]$. The training data and physical constraints are sampled uniformly in both temporal and spatial dimensions. The amount of training data is $N_T = 21 \times 6 \times 6 = 756$, which means that there are 21 sampling points in the temporal dimension, 6 sampling points in the $x$-direction, and 6 in the $y$-direction of the spatial domain. In other words, the grid spacing is $\Delta x = 0.2$ and the time step is $\Delta t = 0.05$ in the FEM solution. The number of physical constraints is $21 \times 11 \times 11 = 2541$, where the grid spacing is $\Delta x = 0.1$ and the time step is $\Delta t = 0.05$ for physical constraints. The numbers of sampling points corresponding to the physical loss, initial loss, and boundary loss are $N_P = 1620$, $N_I = 121$, and $N_S = 800$ respectively, which sum up to 2541. Once the training is finished, the temperature at $t = 1$ will be predicted from different models with a grid spacing of $\Delta x = 0.04$, which is finer than the grid spacings of the training data and physical constraints.

Both GDA and Dual-Dimer methods are used to search high-order saddle points for the PCNN-MM formulation. The gradient descent ascent steps in the GDA and Dual-Dimer method are given by the Adam algorithm with the learning rate of $5 \times 10^{-4}$. The same Adam algorithm with the learning rate of $5 \times 10^{-4}$ is used to minimize the total loss during the training of a PCNN. The dimer distance is $2\Delta R = 2 \times 10^{-4}$. The hyperparameters for the Dual-Dimer method are listed in Table 5. In the heat transfer example, the search for a saddle point stops when the total loss is less than the threshold ($E < \varepsilon$). This is because that the total loss could still be large when the norm of the force is small in the heat transfer example. In the heat transfer example, if the true solution $u$ is found, then the total loss $E$ becomes zero. That is the reason that $E < \varepsilon$ is used as the criteria to determine whether a good prediction to approximate the true solution is found.

Table 5. Hyperparameters of the Dual-Dimer method in the heat transfer example

| Hyperparameters | Value |
|---|---|
| Frequency of updating extreme eigenvalues and eigenvectors, $m$ | 40 |
| The parameter to avoid the zero-division error, $\delta$ | $1 \times 10^{-3}$ |
| Maximum step length of $\Delta\boldsymbol{\theta}_s$ and $\Delta\boldsymbol{\theta}_l$, $\gamma$ | $1 \times 10^{-5}$ |
| Learning rate for the gradient descent ascent sub-steps, $\eta$ | $5 \times 10^{-4}$ |
| The threshold for stopping search ($E < \varepsilon$), $\varepsilon$ | $1 \times 10^{-3}$ |

## 5.2 Computational Results

The predicted temperature fields from different models at $t = 1$ are shown in Fig. 2. The dots in the figures represent the evaluation positions of the temperature field in the 2D domain, where a total of 26×26 samples are taken. It is observed that the predicted temperature fields from the PCNN and PCNN-MMs are close to the FEM solution.

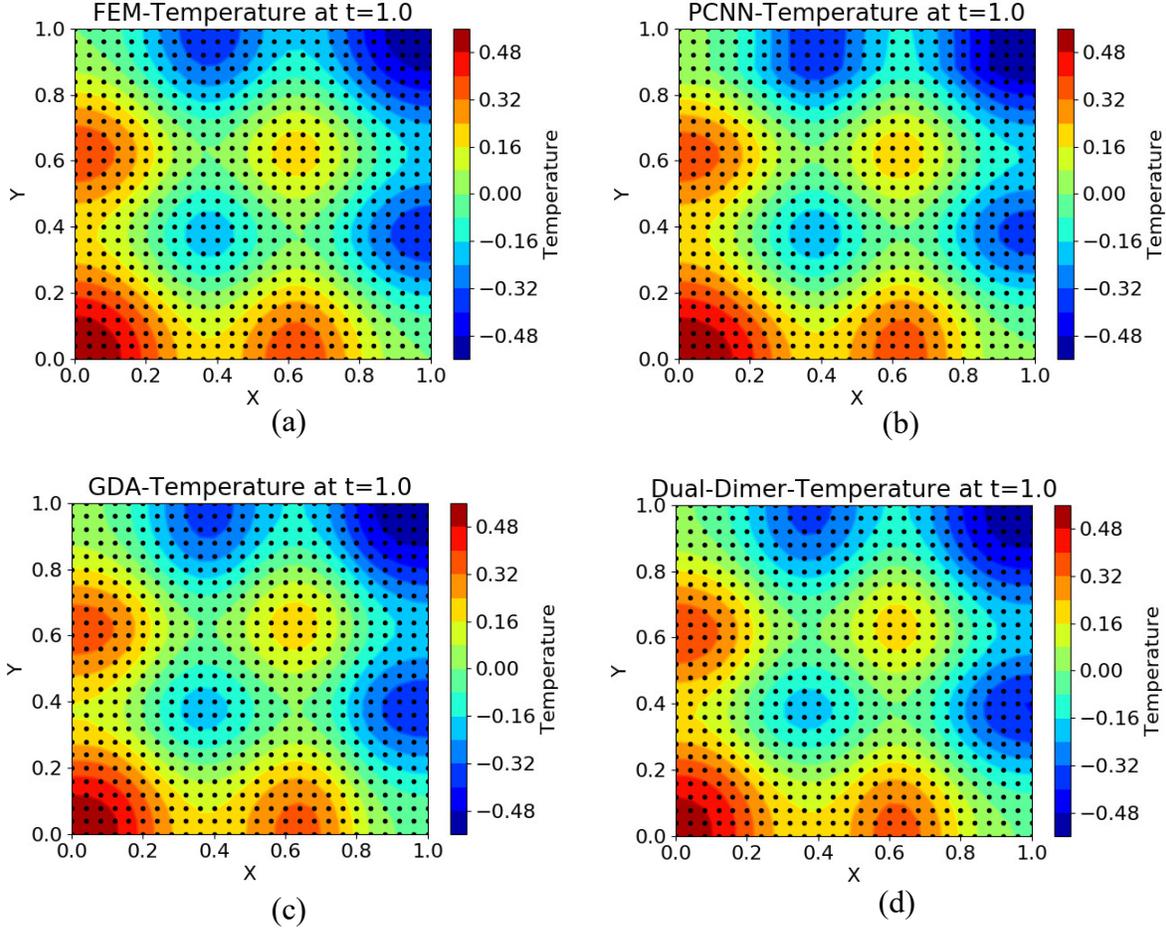

Fig. 2. The predicted temperature fields from different models at $t = 1$: (a) the original FEM solution, (b) the PCNN with the adaptive weighting scheme, (c) the PCNN-MM trained by the GDA method, and (d) the PCNN-MM trained by the Dual-Dimer method.

The changes in losses and weights for different models during the training process are shown in Fig. 3. In general, most losses for different models monotonically decrease during the training. The total loss is less than the desired threshold at the end of the training. However, the convergence speeds of PCNN-MMs are greater than that of the PCNN because the problem formulations are different. The training of the PCNN is to solve the minimization problem, whereas the training of the PCNN-MM is to solve the minimax problem. Note that in the training of the PCNN and PCNN-MMs, the relative importance of the training data and prior knowledge in the total loss function is adjusted dynamically by changing the weights of different losses. As shown in Fig. 3(c), the weights of the PCNN are adjusted dynamically based on the percentages of individual losses in the total loss function. Therefore, a larger weight will be assigned to a larger loss term. As shown in Fig. 3(a), different losses converge at the same speed in the later training stage of the PCNN when different losses have the same magnitude. In the training of PCNN-MMs, the weights of different losses are adjusted dynamically to maximize the total loss. Similarly, a larger weight is assigned to a larger loss term. As shown in Fig. 3(b) and Fig. 3(d), the initial loss is high, whereas the physical loss is low in the early training stage of the PCNN-MM. Therefore, the weight of the initial loss increases, whereas the weight of the physical loss decreases. By minimizing the possible maximum total loss, the convergence speed of the PCNN-MM increases. The changes in losses and weights for different PCNN-MMs are similar because the maximum step lengths of $\Delta\boldsymbol{\theta}_s$ and $\Delta\boldsymbol{\theta}_l$ are small to avoid divergence. By using the information of extreme eigenvalues, the convergence speed of the PCNN-MM trained by the Dual-Dimer method is slightly higher than that of the PCNN-MM trained by the GDA method. Note that the purpose of using the extreme eigenvalues and eigenvectors in the Dual-Dimer method is not to accelerate the convergence, but to make sure that the high-order saddle points are found.

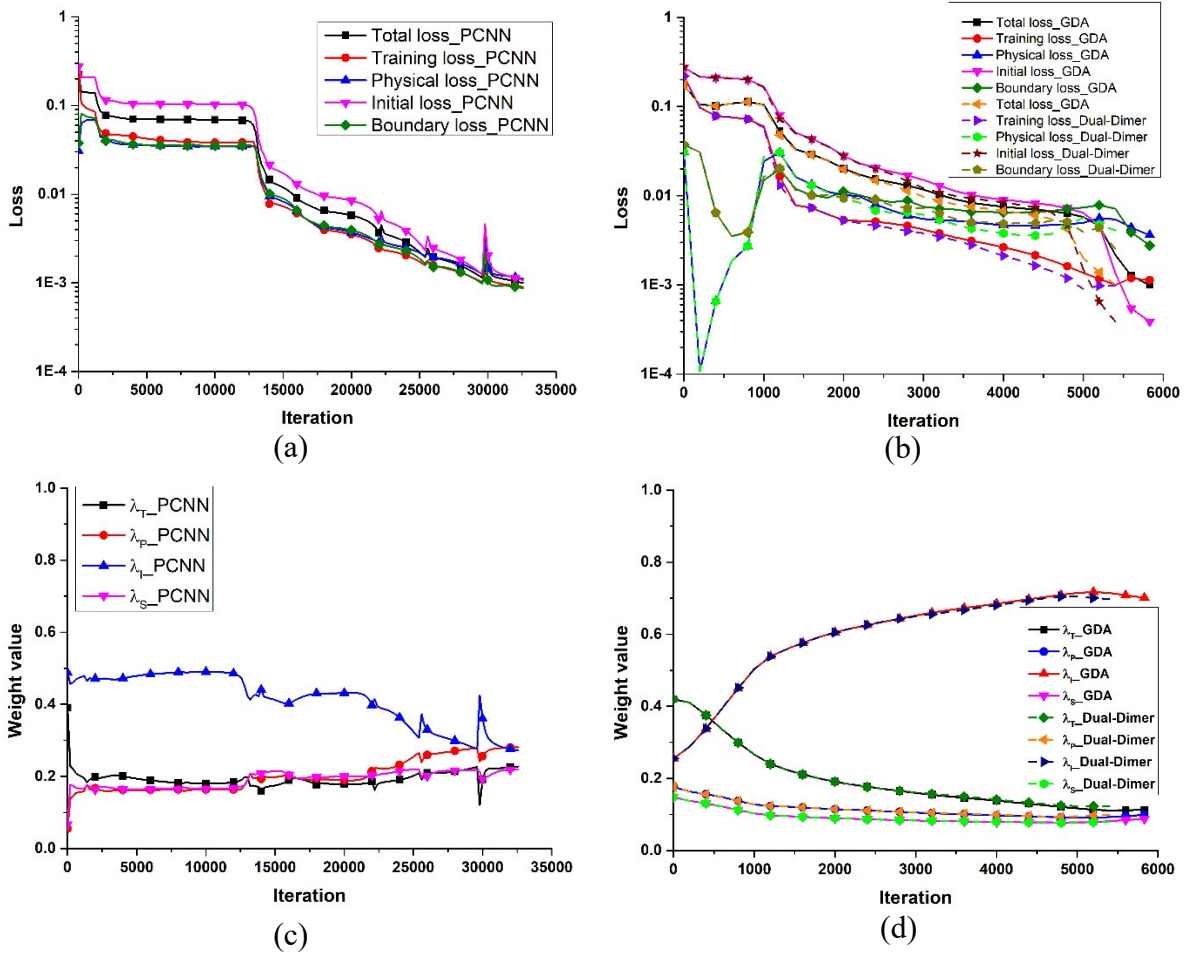

Fig. 3. The changes in losses and weights for different models during the training process: (a) losses of the PCNN, (b) losses of PCNN-MMs, (c) weights of the PCNN, and (d) weights of PCNN-MMs.

The changes in the forces and eigenvalues during the training of PCNN-MMs are shown in Fig. 4. As is shown in Fig. 4(a), the total loss can still be large when the norm of the force is small during the training process. That is the reason that $E < \varepsilon$ is used as the criteria to determine whether a good prediction is found. At the end of the training, the forces for both PCNN-MMs are close to zero, meaning that a critical point is found. Note that eigenvalues are not directly provided by the GDA method. The eigenvalues shown in Fig. 4(b) and Fig. 4(c) are recalculated by the Dual-Dimer method. At the end of the training, the minimum eigenvalue $\beta_s$ in the **w** subspace is positive and maximum eigenvalue $\beta_l$ in the **α** subspace is close to zero. This means that the desired high-order saddle point is found. The results demonstrate the effectiveness of the proposed Dual-Dimer method.

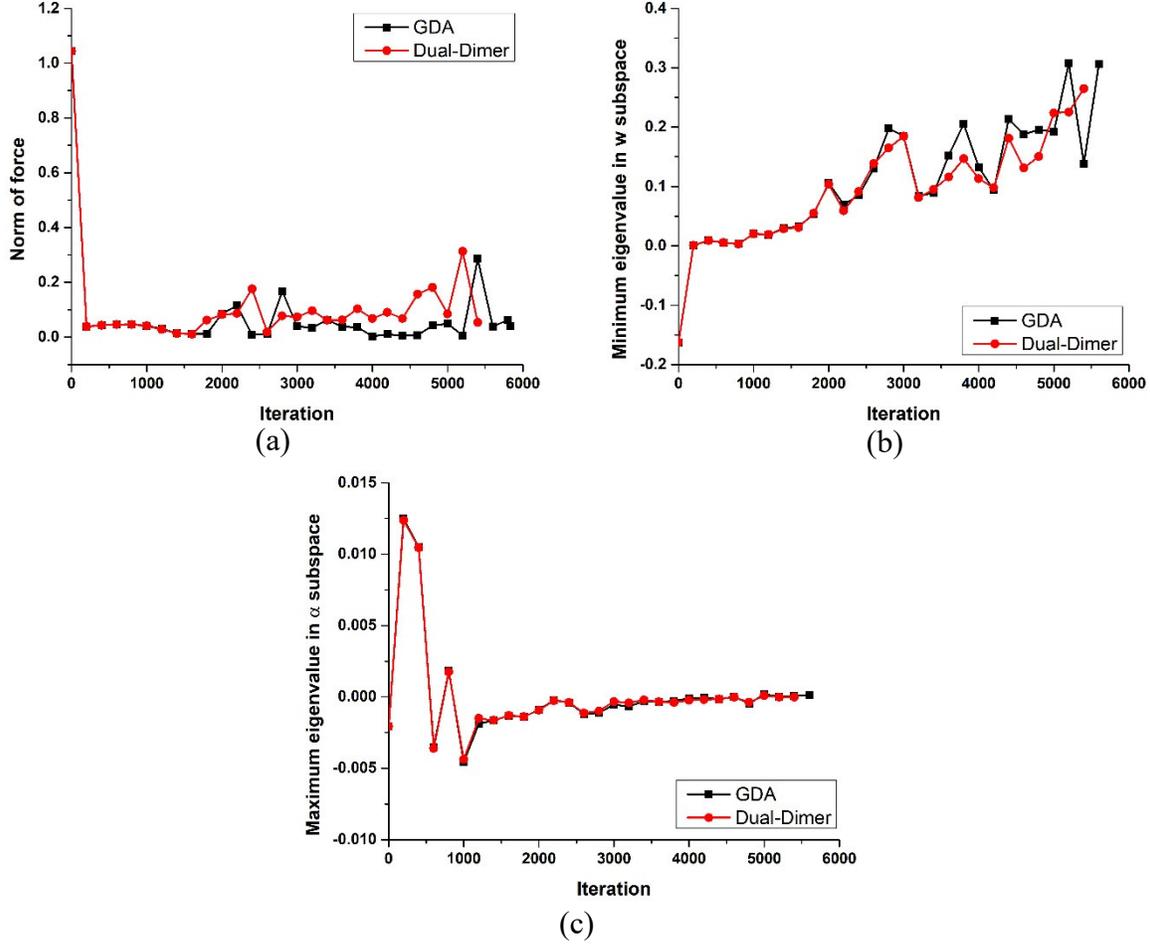

Fig. 4. Forces and eigenvalues during the training of PCNN-MMs: (a) norm of force, (b) minimum eigenvalue $\beta_s$ in the **w** subspace, and (c) maximum eigenvalue $\beta_l$ in the **α** subspace.

To test the convergence speed and stability of different models, the PCNN and PCNN-MMs were run 20 times with random initial weights of neural networks. The mean values of training iterations, training time, and mean squared error (MSE) for different models are shown in Table 6, where their standard deviations are also shown in parentheses. Fig. 5 shows that the convergence speeds of PCNN-MMs are about 3 times faster than that of the PCNN, whereas the MSEs of predictions by PCNN-MMs at $t = 1$ are slightly larger than that by the PCNN. The MSEs of predictions by the PCNN and PCNN-MMs are all less than the error threshold $\varepsilon = 1 \times 10^{-3}$ with negligible differences. The results show the increased computational efficiency of PCNNs by adopting the new minimax architecture. The standard deviations of the training iterations and training time by PCNN-MMs are less than those by the PCNN, whereas the standard deviations of the MSEs of prediction by PCNN-MMs at $t = 1$ are slightly larger than that by the PCNN. The results also indicate the stability of the proposed PCNN-MMs. The training times of the PCNN-MMs by the GDA method and the Dual-Dimer method are similar. However, the Dual-Dimer method can provide additional eigenvalue information to make sure that the desired high-order saddle points are found at the end of the training.

The above computational results demonstrate that PCNN-MMs are computationally more efficient in training than the original PCNN with adaptive weighting scheme. The proposed minimax architecture has the advantage of systematically adjusting the weights of different losses. The results also show that the local convergence of PCNN-MMs is stable. In addition, with the similar accuracy and efficiency of the GDA method, the Dual-Dimer method can provide additional eigenvalue information to make sure that the desired saddle points are found at the end of the training.

Table 6. Quantitative comparison for different models to solve the heat transfer problem

| Models | Training iterations | Training time (second) | MSE of prediction at $t = 1$ | Minimum eigenvalue $\beta_s$ in the $\mathbf{w}$ subspace at the end of the training | Maximum eigenvalue $\beta_l$ in the $\boldsymbol{\alpha}$ subspace at the end of the training |
|---|---|---|---|---|---|
| PCNN with the adaptive weighting scheme | 58497 (24878) | 2259.46 (930.81) | $3.24 \times 10^{-4}$ $(1.62 \times 10^{-4})$ | N/A | N/A |
| PCNN-MM with the GDA method | 15322 (7023) | 614.72 (247.48) | $4.22 \times 10^{-4}$ $(3.72 \times 10^{-4})$ | 0.95 (0.78) | $5.84 \times 10^{-5}$ $(8.19 \times 10^{-5})$ |
| PCNN-MM with the Dual-Dimer method | **13376 (6035)** | **560.85 (246.08)** | $5.56 \times 10^{-4}$ $(4.13 \times 10^{-4})$ | 0.71 (0.53) | $-6.91 \times 10^{-5}$ $(1.77 \times 10^{-4})$ |

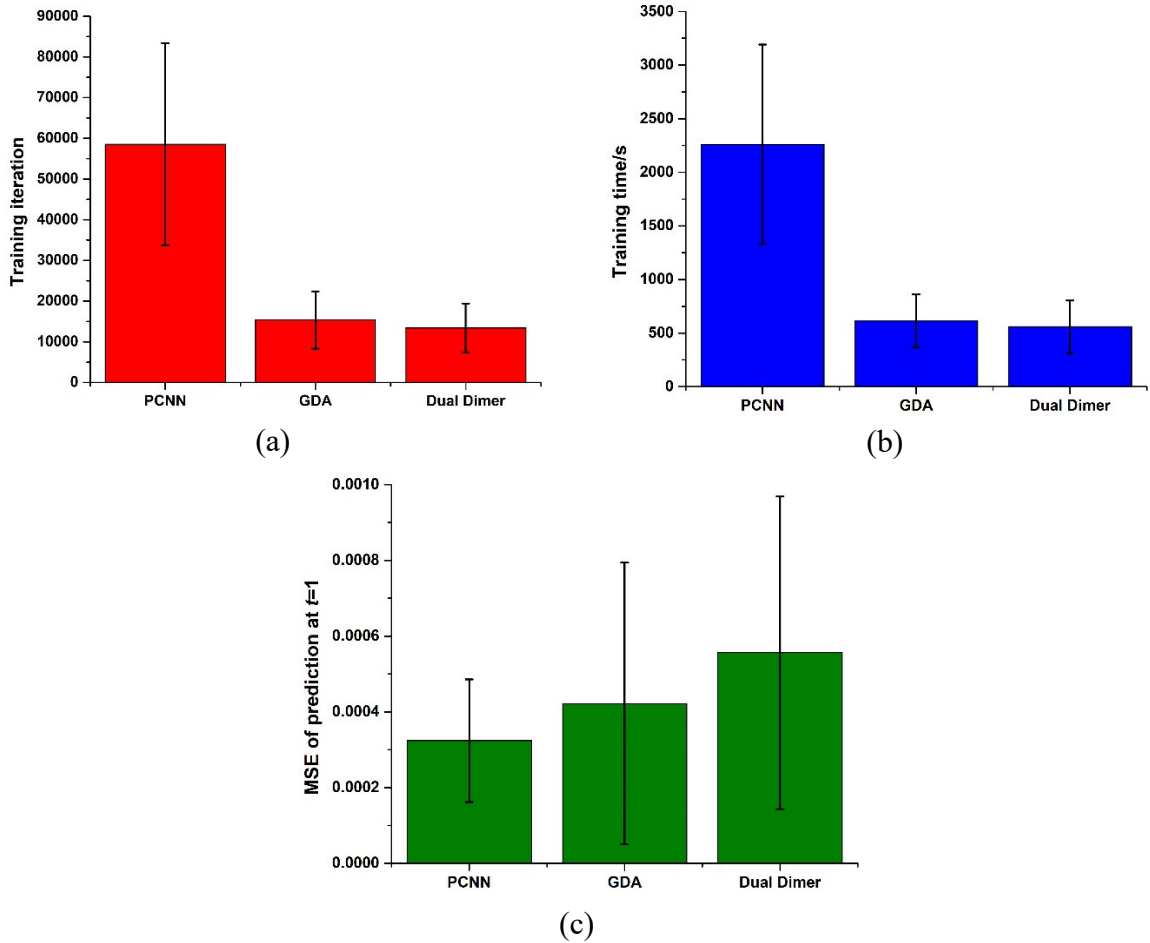

Fig. 5. Quantitative comparison for different models in (a) training iteration, (b) training time, and (c) MSE of prediction at $t = 1$.

## 6   CONCLUSIONS

In this work, a new physics-constrained neural network with the minimax architecture is proposed to adjust the weights of different losses systematically. The training of the PCNN-MM is to solve a minimax problem and search for the high-order saddle points of the nonconvex-nonconcave loss function. To address the challenges of searching high-order saddle points, a novel saddle point search algorithm called Dual-Dimer method is proposed, where only first derivatives need to be calculated. The local convergence of the Dual-Dimer method is analyzed. The performance of the Dual-Dimer method is evaluated with three analytical nonconvex-nonconcave loss functions. It was shown that the Dual-Dimer method is computationally more efficient than the GDA method to find high-order saddle points in these analytical functions. The Dual-Dimer method also provides additional eigenvalue information to make sure that the desired high-order saddle points are found at the end of the training. A heat transfer example is used to demonstrate the effectiveness of the PCNN-MM, where its convergence is faster than that of the original PCNN with adaptive weighting scheme.

The adjustment of hyperparameters in this study is based on sensitivity studies. In future work, a more systematic method to find the optimal hyperparameters will be developed so that the computational efficiency of the Dual-Dimer method can be further improved. In addition, using more eigenvalues and eigenvectors in the Dual-Dimer method can potentially accelerate the saddle point search. Further investigation is needed. The generic Dual-Dimer method can be applied to solve other minimax problems, which arise from game theory, generative adversarial networks, and robust optimization.

## ACKNOWLEDGEMENT

This work was supported in part by the George W. Woodruff Faculty Fellowship at Georgia Institute of Technology.

**APPENDIX**

**Lemma 1.** The Jacobian of the loss function at the desired saddle point $\boldsymbol{\theta}^* = (\mathbf{w}^*, \boldsymbol{\alpha}^*)$ is

$$\nabla F(\boldsymbol{\theta}^*) = \mathbf{I} + \eta \begin{pmatrix} -\nabla_{\mathbf{w}}^2 E(\boldsymbol{\theta}^*) & -\nabla_{\mathbf{w},\boldsymbol{\alpha}}^2 E(\boldsymbol{\theta}^*) \\ \nabla_{\boldsymbol{\alpha},\mathbf{w}}^2 E(\boldsymbol{\theta}^*) & \nabla_{\boldsymbol{\alpha}}^2 E(\boldsymbol{\theta}^*) \end{pmatrix} + \begin{pmatrix} -\frac{1}{\beta_s}\mathbf{v}_s\mathbf{v}_s^T\nabla_{\mathbf{w}}^2 E(\boldsymbol{\theta}^*) & -\frac{1}{\beta_s}\mathbf{v}_s\mathbf{v}_s^T\nabla_{\mathbf{w},\boldsymbol{\alpha}}^2 E(\boldsymbol{\theta}^*) \\ -\frac{1}{\beta_l}\mathbf{v}_l\mathbf{v}_l^T\nabla_{\boldsymbol{\alpha},\mathbf{w}}^2 E(\boldsymbol{\theta}^*) & -\frac{1}{\beta_l}\mathbf{v}_l\mathbf{v}_l^T\nabla_{\boldsymbol{\alpha}}^2 E(\boldsymbol{\theta}^*) \end{pmatrix},$$

where $\mathbf{I}$ is the real-valued identity matrix. Furthermore, if there is an $\eta > 0$ such that the absolute values of all eigenvalues of $\nabla F(\boldsymbol{\theta}^*)$ are less than 1, then there is an open neighborhood $K$ of $\boldsymbol{\theta}^*$ so that for all $\boldsymbol{\theta} \in K$, the fixed-point iterations of $F(\boldsymbol{\theta})$ in Eq. (15) are stable in $K$. The rate of convergence is at least linear.

*Proof.* Since $\boldsymbol{\theta}^*$ is a desired saddle point, we have $\nabla_{\mathbf{w}}E(\boldsymbol{\theta}^*) = \mathbf{0}$, $\nabla_{\boldsymbol{\alpha}}E(\boldsymbol{\theta}^*) = \mathbf{0}$, $\beta_s \geq 0$, and $\beta_l \leq 0$. Furthermore $F(\boldsymbol{\theta}^*) = \boldsymbol{\theta}^*$. The Jacobian $\nabla F(\boldsymbol{\theta}^*)$ is given by

$$\nabla F(\boldsymbol{\theta}^*) = \mathbf{I} + \eta \begin{pmatrix} -\nabla_{\mathbf{w}}^2 E(\boldsymbol{\theta}^*) & -\nabla_{\mathbf{w},\boldsymbol{\alpha}}^2 E(\boldsymbol{\theta}^*) \\ \nabla_{\boldsymbol{\alpha},\mathbf{w}}^2 E(\boldsymbol{\theta}^*) & \nabla_{\boldsymbol{\alpha}}^2 E(\boldsymbol{\theta}^*) \end{pmatrix} + \nabla\left(-\frac{(\mathbf{v}_s\cdot\nabla_{\mathbf{w}}E(\boldsymbol{\theta}^*))\mathbf{v}_s}{|\beta_s|}, \frac{(\mathbf{v}_l\cdot\nabla_{\boldsymbol{\alpha}}E(\boldsymbol{\theta}^*))\mathbf{v}_l}{|\beta_l|}\right).$$

If $\beta_s > 0$, then we have

$$\begin{aligned}
\nabla\left(-\frac{(\mathbf{v}_s\cdot\nabla_{\mathbf{w}}E(\boldsymbol{\theta}^*))\mathbf{v}_s}{|\beta_s|}\right) &= \nabla\left(-\frac{(\mathbf{v}_s\cdot\nabla_{\mathbf{w}}E(\boldsymbol{\theta}^*))\mathbf{v}_s}{\beta_s}\right) = -\mathbf{v}_s \otimes \nabla\left(\frac{\mathbf{v}_s\cdot\nabla_{\mathbf{w}}E(\boldsymbol{\theta}^*)}{\beta_s}\right) - \left(\frac{\mathbf{v}_s\cdot\overbrace{\nabla_{\mathbf{w}}E(\boldsymbol{\theta}^*)}^{\mathbf{0}}}{\beta_s}\right)\nabla\mathbf{v}_s \\
&= -\mathbf{v}_s \otimes \left[\frac{1}{\beta_s}\nabla(\mathbf{v}_s\cdot\nabla_{\mathbf{w}}E(\boldsymbol{\theta}^*)) + \left(\mathbf{v}_s\cdot\overbrace{\nabla_{\mathbf{w}}E(\boldsymbol{\theta}^*)}^{\mathbf{0}}\right)\nabla\frac{1}{\beta_s}\right] \\
&= -\frac{1}{\beta_s}\mathbf{v}_s \otimes \left[\nabla^T\mathbf{v}_s\overbrace{\nabla_{\mathbf{w}}E(\boldsymbol{\theta}^*)}^{\mathbf{0}} + \nabla^T(\nabla_{\mathbf{w}}E(\boldsymbol{\theta}^*))\mathbf{v}_s\right] \\
&= -\frac{1}{\beta_s}\mathbf{v}_s\mathbf{v}_s^T\nabla(\nabla_{\mathbf{w}}E(\boldsymbol{\theta}^*)) \\
&= \left(-\frac{1}{\beta_s}\mathbf{v}_s\mathbf{v}_s^T\nabla_{\mathbf{w}}^2 E(\boldsymbol{\theta}^*) \quad -\frac{1}{\beta_s}\mathbf{v}_s\mathbf{v}_s^T\nabla_{\mathbf{w},\boldsymbol{\alpha}}^2 E(\boldsymbol{\theta}^*)\right)
\end{aligned} \tag{29}$$

Similarly, if $\beta_l < 0$, we have

$$\nabla\left(\frac{(\mathbf{v}_l\cdot\nabla_{\boldsymbol{\alpha}}E(\boldsymbol{\theta}^*))\mathbf{v}_l}{|\beta_l|}\right) = \left(-\frac{1}{\beta_l}\mathbf{v}_l\mathbf{v}_l^T\nabla_{\boldsymbol{\alpha},\mathbf{w}}^2 E(\boldsymbol{\theta}^*) \quad -\frac{1}{\beta_l}\mathbf{v}_l\mathbf{v}_l^T\nabla_{\boldsymbol{\alpha}}^2 E(\boldsymbol{\theta}^*)\right). \tag{30}$$

Therefore, we have the Jacobian

$$\nabla F(\boldsymbol{\theta}^*) = \mathbf{I} + \eta \begin{pmatrix} -\nabla_{\mathbf{w}}^2 E(\boldsymbol{\theta}^*) & -\nabla_{\mathbf{w},\boldsymbol{\alpha}}^2 E(\boldsymbol{\theta}^*) \\ \nabla_{\boldsymbol{\alpha},\mathbf{w}}^2 E(\boldsymbol{\theta}^*) & \nabla_{\boldsymbol{\alpha}}^2 E(\boldsymbol{\theta}^*) \end{pmatrix} + \begin{pmatrix} -\frac{1}{\beta_s} \mathbf{v}_s \mathbf{v}_s^T \nabla_{\mathbf{w}}^2 E(\boldsymbol{\theta}^*) & -\frac{1}{\beta_s} \mathbf{v}_s \mathbf{v}_s^T \nabla_{\mathbf{w},\boldsymbol{\alpha}}^2 E(\boldsymbol{\theta}^*) \\ -\frac{1}{\beta_l} \mathbf{v}_l \mathbf{v}_l^T \nabla_{\boldsymbol{\alpha},\mathbf{w}}^2 E(\boldsymbol{\theta}^*) & -\frac{1}{\beta_l} \mathbf{v}_l \mathbf{v}_l^T \nabla_{\boldsymbol{\alpha}}^2 E(\boldsymbol{\theta}^*) \end{pmatrix}.$$

According to the fixed point theorem (Mescheder et al., 2017), if there is an $\eta > 0$ such that the absolute values of the eigenvalues of the Jacobian $\nabla F(\boldsymbol{\theta}^*)$ are all smaller than 1, then there is an open neighborhood $K$ of $\boldsymbol{\theta}^*$ so that for all $\boldsymbol{\theta} \in K$, the iterates of $F(\boldsymbol{\theta})$ in Eq. (15) are stable. The rate of convergence is at least linear.

**Lemma 2.** Let $\beta_A = a + bi$ be the eigenvalues of the matrix $\mathbf{A}$, $\beta_B = c + di$ be the eigenvalues of the matrix $\mathbf{B}$, where $i = \sqrt{-1}$. The eigenvalues of the matrix $\mathbf{I} + \eta \mathbf{A} + \mathbf{B}$, where $\eta > 0$, lie in the unit ball if

$$\Delta = [2(a + ac + bd)]^2 - 4(a^2 + b^2)(c^2 + 2c + d^2) > 0$$

and

$$\begin{cases} 0 < \eta < \frac{-2(a+ac+bd)+\sqrt{\Delta}}{2(a^2+b^2)}, \ if \ a + ac + bd \geq 0 \ and \ c^2 + 2c + d^2 > 0 \\ max\left\{0, \frac{-2(a+ac+bd)-\sqrt{\Delta}}{2(a^2+b^2)}\right\} < \eta < \frac{-2(a+ac+bd)+\sqrt{\Delta}}{2(a^2+b^2)}, \ if \ a + ac + bd < 0 \end{cases}$$

for all eigenvalues $\beta_A$ of $\mathbf{A}$ and $\beta_B$ of $\mathbf{B}$.

*Proof.* If the eigenvalues of the matrix $\mathbf{I} + \eta \mathbf{A} + \mathbf{B}$ lie in the unit ball, then $|1 + \eta\beta_A + \beta_B|^2 < 1$. That is, $(1 + \eta a + b)^2 + (\eta c + d)^2 < 1$, which leads to

$$(a^2 + b^2)\eta^2 + 2(a + ac + bd)\eta + c^2 + 2c + d^2 < 0. \tag{31}$$

To find the real solutions of $\eta$, we need to make sure the discriminant is larger than zero, as

$$\Delta = [2(a + ac + bd)]^2 - 4(a^2 + b^2)(c^2 + 2c + d^2) > 0. \tag{32}$$

Two real roots $\eta_1 = \frac{-2(a+ac+bd)-\sqrt{\Delta}}{2(a^2+b^2)}$ and $\eta_2 = \frac{-2(a+ac+bd)+\sqrt{\Delta}}{2(a^2+b^2)}$ can be obtained. Since $\eta > 0$, we have

$$\eta_2 = \frac{-2(a+ac+bd)+\sqrt{\Delta}}{2(a^2+b^2)} > 0 \ or \ \sqrt{\Delta} > 2(a + ac + bd). \tag{33}$$

If $a + ac + bd \geq 0$, then $\Delta > [2(a + ac + bd)]^2$. Therefore,

$$c^2 + 2c + d^2 > 0. \tag{34}$$

Meanwhile, it is obvious that $\eta_1 = \frac{-2(a+ac+bd)-\sqrt{\Delta}}{2(a^2+b^2)} < 0$. Therefore, the range of $\eta$ should be $0 < \eta < \eta_2$ in order to satisfy Eq. (31). If $a + ac + bd < 0$, automatically $\eta_2 = \frac{-2(a+ac+bd)+\sqrt{\Delta}}{2(a^2+b^2)} > 0$ without any further conditions. The range of $\eta$ should be $max\{0, \eta_1\} < \eta < \eta_2$ to satisfy Eq. (31).

**Theorem 1.** Let $\boldsymbol{\theta}^* = (\mathbf{w}^*, \boldsymbol{\alpha}^*)$ be the desired saddle point, $\beta_A = a + bi$ be the eigenvalues of $\mathbf{A} =$

$\begin{pmatrix} -\nabla_{\mathbf{w}}^2 E(\boldsymbol{\theta}^*) & -\nabla_{\mathbf{w},\boldsymbol{\alpha}}^2 E(\boldsymbol{\theta}^*) \\ \nabla_{\boldsymbol{\alpha},\mathbf{w}}^2 E(\boldsymbol{\theta}^*) & \nabla_{\boldsymbol{\alpha}}^2 E(\boldsymbol{\theta}^*) \end{pmatrix}$ , $\beta_B = c + di$ be the eigenvalues of $\mathbf{B} =$

$\begin{pmatrix} -\frac{1}{\beta_s}\mathbf{v}_s\mathbf{v}_s^T\nabla_{\mathbf{w}}^2 E(\boldsymbol{\theta}^*) & -\frac{1}{\beta_s}\mathbf{v}_s\mathbf{v}_s^T\nabla_{\mathbf{w},\boldsymbol{\alpha}}^2 E(\boldsymbol{\theta}^*) \\ -\frac{1}{\beta_l}\mathbf{v}_l\mathbf{v}_l^T\nabla_{\boldsymbol{\alpha},\mathbf{w}}^2 E(\boldsymbol{\theta}^*) & -\frac{1}{\beta_l}\mathbf{v}_l\mathbf{v}_l^T\nabla_{\boldsymbol{\alpha}}^2 E(\boldsymbol{\theta}^*) \end{pmatrix}$, and $\eta > 0$. The fixed-point iterations of $F(\boldsymbol{\theta})$ in Eq. (15)

are locally stable if

$$\Delta = [2(a + ac + bd)]^2 - 4(a^2 + b^2)(c^2 + 2c + d^2) > 0$$

and

$$\begin{cases} 0 < \eta < \frac{-2(a+ac+bd)+\sqrt{\Delta}}{2(a^2+b^2)}, \ if \ a + ac + bd \geq 0 \ and \ c^2 + 2c + d^2 > 0 \\ max\left\{0, \frac{-2(a+ac+bd)-\sqrt{\Delta}}{2(a^2+b^2)}\right\} < \eta < \frac{-2(a+ac+bd)+\sqrt{\Delta}}{2(a^2+b^2)}, \ if \ a + ac + bd < 0 \end{cases}$$

for all eigenvalues $\beta_A$ of $\mathbf{A}$ and $\beta_B$ of $\mathbf{B}$.

*Proof.* This is a direct consequence of Lemma 1 and Lemma 2.